\documentclass[10pt,twocolumn,letterpaper]{article}

\usepackage{iccv}              % To produce the CAMERA-READY version
\definecolor{iccvblue}{rgb}{0.21,0.49,0.74}
\usepackage[pagebackref,breaklinks,colorlinks,allcolors=iccvblue]{hyperref}

\usepackage{textcomp}
\usepackage{stfloats}
\usepackage{url}
\usepackage{verbatim}
\usepackage{graphicx}

\usepackage{times}
\usepackage{epsfig}
\usepackage{graphicx}
\usepackage{amssymb}
\usepackage{mathabx}
\usepackage{multirow}
\usepackage{fontawesome}
\usepackage{multicol}

\usepackage{color}
\usepackage{pifont}

\def\paperID{11013} % *** Enter the Paper ID here
\def\confName{ICCV}
\def\confYear{2025}

\title{Towards Differential Handling of Various Blur Regions for Accurate Image Deblurring}

\author{Hu Gao\\
Beijing Normal University\\
Beijing, China\\
{\tt\small gao\_h@mail.bnu.edu.cn}
\and
Depeng Dang\\
Beijing Normal University\\
Beijing, China\\
{\tt\small ddepeng@bnu.edu.cn}
}

\begin{document}
\maketitle
\begin{abstract}
Image deblurring aims to restore high-quality images by removing undesired degradation. Although existing methods have yielded promising results, they either overlook the varying degrees of degradation across different regions of the blurred image. In this paper, we propose a differential handling network (DHNet) to perform differential processing for different blur regions. Specifically, we design a Volterra block (VBlock) to incorporate nonlinear characteristics into the deblurring network, enabling it to map complex input-output relationships without relying on nonlinear activation functions. 
To enable the model to adaptively address varying degradation degrees in blurred regions, we devise the degradation degree recognition expert module (DDRE). This module initially incorporates prior knowledge from a well-trained model to estimate spatially variable blur information. Consequently, the router can map the learned degradation representation and allocate weights to experts according to both the degree of degradation and the size of the regions.
Comprehensive experimental results show that DHNet effectively surpasses state-of-the-art (SOTA) methods on both synthetic and real-world datasets.
\end{abstract}
    
\section{Introduction}
\label{sec:intro}
Image deblurring seeks to restore high-quality images from these blur versions. Due to the ill-posed nature, traditional method~\cite{karaali2017edge} attempts to tackle it by imposing various priors to limit the solution space. However, creating such priors is difficult and often lacks generalizability, making them impractical for real-world applications.

In recent years, a variety of Convolutional Neural Networks (CNNs)~\cite{FSNet, SFNet, IRNeXt, Zamir2021MPRNet,alggao2024learning} have been developed for image deblurring. The fundamental component of a CNN is a convolutional layer followed by an activation function. The convolution operation ensures local connectivity and translational invariance, while the activation function adds non-linearity to the network. However, CNNs face inherent limitations, such as local receptive fields and a lack of dependence on input content, which restrict their ability to eliminate long-range blur degradation perturbation. To overcome such limitations, Transformers~\cite{kong2023efficient, potlapalli2023promptir, Zamir2021Restormer, u2former} have been introduced in image deblurring. They leverage the adaptive weights of the self-attention mechanism and excel at capturing global dependencies, demonstrating superior performance compared to CNN-based approaches.

\begin{figure}[tb] % use float package if you want it here
	\centering
	\includegraphics[width=1\linewidth]{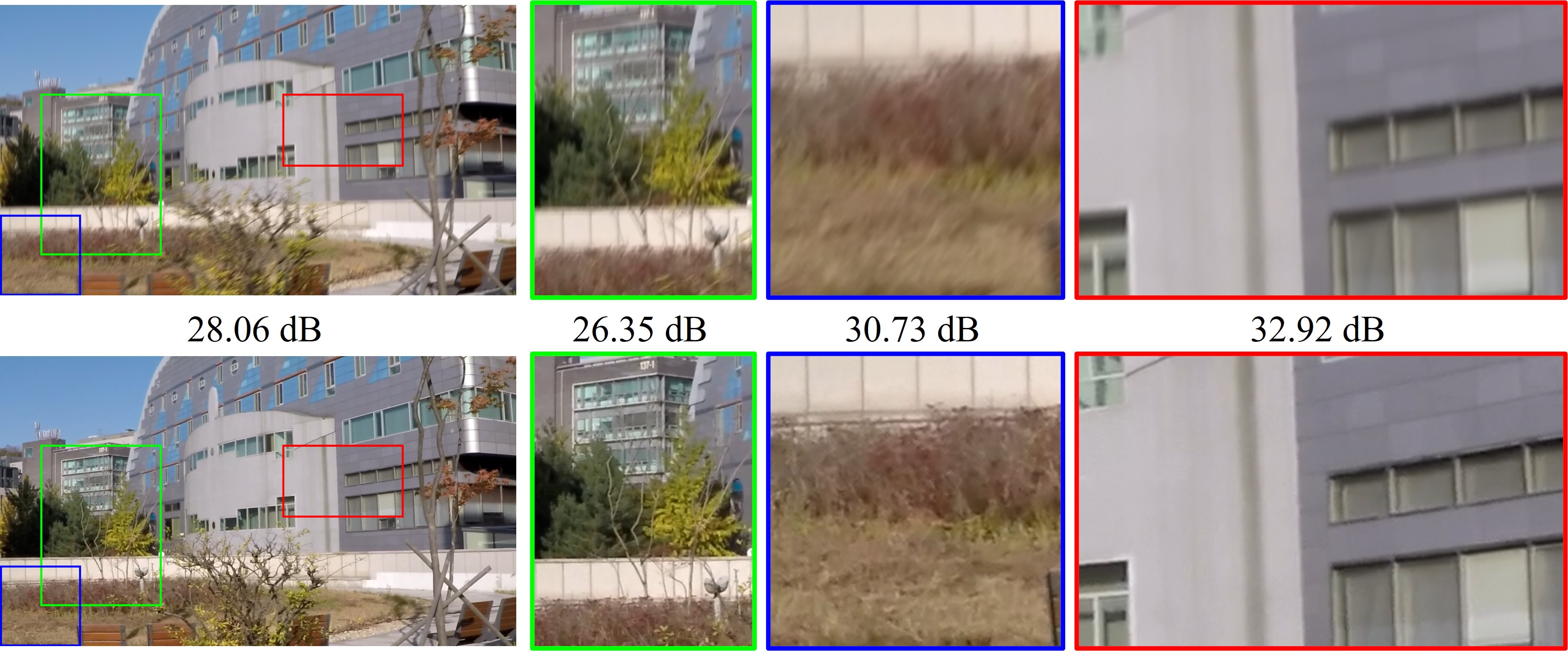}
	\caption{Varying degrees of degradation across different regions. The top row is the blurred image and the bottom row is the sharp image.}
	\label{fig:ques}
\end{figure}
While the aforementioned methods have demonstrated strong performance in image deblurring, they have two key drawbacks: (1) they approximate nonlinear function properties by stacking numerous nonlinear activation functions, and (2) they overlook the varying degrees of degradation across different blur regions of the image. As shown in Figure~\ref{fig:ques}, the degradation in a blurred image varies across different regions. The areas highlighted by the \textcolor{green}{green box} are more severely blurred than those marked by the \textcolor{red}{red box}.

To tackle the first drawback, NAFNet~\cite{chen2022simple} argues that high complexity is unnecessary and proposes a simpler baseline that employs element-wise multiplication instead of nonlinear activation functions, achieving better results with lower computational resources. However, this approach inevitably sacrifices the ability to map complex input-output relationships, making it difficult to manage more intricate scenes. To address the second issue, AdaRevD~\cite{AdaRevD} introduces a classifier to assess the degradation degree of image patches. However, this method relies on a limited number of predefined categories and a fixed blur patch size\footnote{AdaRevD groups the patches into six degradation degrees based on the PSNR between the blurred patch and the sharp patch, and then uses a relatively large patch size of 384 x 384 to classify the degradation degree of each blurred patch.} to specify the degree of degradation, which diminishes its effectiveness in adaptively managing different degrees of degradation across various sizes of blurred patches.
\begin{figure}[tb] % use float package if you want it here
	\centering
	\includegraphics[width=1\linewidth]{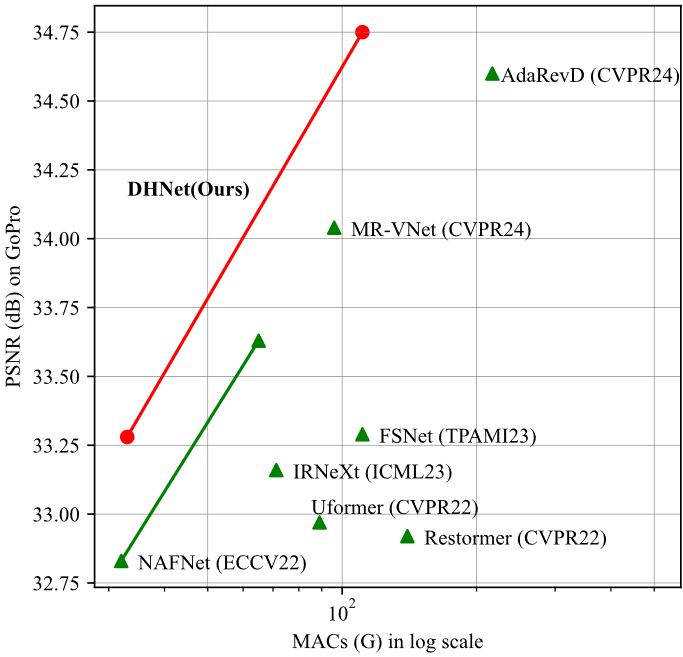}
	\caption{Computational cost vs. PSNR of models on the GoPro dataset~\cite{Gopro}. Our DHNet achieve the SOTA performance with up to 48.9\% of cost reduction.}
	\label{fig:param}
\end{figure}

Based on the analyses presented above, a natural question arises: Is it feasible to devise a network that effectively  performs differential processing for different blur regions with less nonlinear activation function? In pursuit of this objective, we propose DHNet for efficient image deblurring, incorporating several key components. 
1)  We design a Volterra block (VBlock) to investigate non-linearity within the network. VBlock avoids using traditional nonlinear activation functions and instead employs Volterra kernel to enhance linear convolution by facilitating interactions between image pixels. This approach approximates non-linearity without the computational overhead associated with stacking numerous nonlinear functions to map complex input-output relationships.
2) To adaptively identify degradation degrees across varying sizes of blur regions, we propose a degradation degree  recognition expert module (DDRE). DDRE first integrates prior knowledge from a well-trained model to estimate the spatially variable blur information. This enables the router to map the learned degradation representation and assign weights to experts based on both the degree of degradation and the size of the regions. 
We conduct extensive experiments to validate the effectiveness of the proposed networks, demonstrating their remarkable performance advantage over state-of-the-art approaches. As illustrated in Figure~\ref{fig:param}, our DHNet model achieves SOTA performance while requiring less computational cost compared to existing methods.

The main contributions are summarized as follows:
\begin{enumerate}
	\item  We propose an efficient and effective framework for image deblurring, called DHNet, which excels at differentially handling various blur regions while maintaining lower computational costs.
 
    \item  We design a Volterra block (VBlock) to investigate non-linearity within the network, avoiding the previous operation of stacking numerous nonlinear functions to map complex input-output relationships. 
    
    \item  We devise a degradation degree recognition expert module (DDRE), enabling the model adaptively deal with the different degradation degrees of the degraded region.
    
    \item Extensive experiments demonstrate that the proposed DHNet achieves promising performance compared to state-of-the-art methods across synthetic and real-world benchmark datasets.
\end{enumerate}
\section{Related Work}
\label{sec:formatting}

\subsection{Image Deblurring}
The ill-posed nature of image deblurring leads many conventional approaches~\cite{2011Image,karaali2017edge} to rely on hand-crafted priors to limit the possible solutions. Although these priors can assist in blur removal, they often struggle to accurately model the degradation process and often lack generalization.

Rather than manually designing image priors, many methods focus on developing various deep CNNs~\cite{AdaRevD,Zamir2021MPRNet,Zamir2022MIRNetv2,FSNet,SFNet,focalnetcui2023focal,IRNeXt,chen2022simple,CascadedGaze,MR-VNet, DDAnet} to tackle image deblurring.
MPRNet~\cite{Zamir2021MPRNet} presents a multi-stage progressive approach that improves the exploration of spatial details and contextual information. 
MIRNet-V2~\cite{Zamir2022MIRNetv2} employs a multi-scale architecture that maintains spatially precise representations while also capturing complementary contextual insights. 
CGNet~\cite{CascadedGaze} integrates a global context extractor to effectively gather global contextual information. IRNeXt~\cite{IRNeXt} rethinks CNN design and introduces an efficient network. FSNet~\cite{FSNet} utilizes multi-branch and content-aware modules to dynamically select the most informative components. MR-VNet~\cite{MR-VNet} proposes a novel architecture that leverages Volterra layers for both image and video restoration. 
Nonetheless, the inherent characteristics of convolutional operations, such as local receptive fields, limit the models' ability to effectively address long-range degradation disturbances.

To tackle these limitations, Transformers~\cite{kong2023efficient, potlapalli2023promptir, DeblurDiNAT, Zamir2021Restormer, MRLPFNet, Wang_2022_CVPR, mt10387581, liang2021swinir, Tsai2022Stripformer} have been utilized in image deblurring. They effectively capture global dependencies through the adaptive weights of the self-attention mechanism, outperforming CNN-based methods. However, the quadratic time complexity of the self-attention mechanism with respect to input size increases the computational burden.
To mitigate this issue, Uformer~\cite{Wang_2022_CVPR}, SwinIR~\cite{liang2021swinir}, and U$^2$former~\cite{u2former} implement self-attention using a window-based approach. In contrast, Restormer~\cite{Zamir2021Restormer}, MRLPFNet~\cite{MRLPFNet}, and DeblurDiNAT~\cite{DeblurDiNAT} compute self-attention across channels rather than in the spatial dimension, achieving linear complexity in relation to input size. Additionally, FFTformer~\cite{kong2023efficient} leverages frequency domain properties to estimate scaled dot-product attention.

While the aforementioned methods have two main drawbacks: (1) they approximate nonlinear functions by stacking many nonlinear activation functions, and (2) they fail to account for the varying degrees of degradation in different blur regions.
Although NAFNet~\cite{chen2022simple} uses element-wise multiplication instead of nonlinear activation functions, it struggles with more complex scenes. AdaRevD~\cite{AdaRevD} introduces a classifier to assess the degradation degree of image patches, but it relies on a limited number of predefined categories and a fixed patch size, lacking adaptive flexibility.  
In this paper, we propose a differential handling network that incorporates two key components: the Volterra block (VBlock) and the degradation degree recognition expert module (DDRE). This network achieves nonlinear properties with fewer nonlinear activation functions while adaptively handling varying degradation degrees in blur regions of different sizes.

\subsection{Volterra Series}
The Volterra series is a model for nonlinear behavior that effectively captures "memory" effects~\cite{Volterra2005TheoryOF}. With its capability to learn nonlinear functions, Volterra Filters have been applied in deep learning~\cite{MR-VNet, Volter9247263, Volter8237772, roheda2024volterra}. To enhance non-linearity and complement traditional activation functions,~\cite{Volter8237772} introduces a single layer of Volterra kernel-based convolutions followed by standard CNN layers. VNNs~\cite{roheda2024volterra} proposes a cascaded approach of Volterra Filtering to significantly reduce the number of parameters. VolterraNet~\cite{Volter9247263} presents a novel higher-order Volterra convolutional neural network designed for data represented as samples of functions on Riemannian homogeneous spaces. MR-VNet~\cite{MR-VNet} uses Volterra layers to effectively introduce nonlinearities into the image restoration process. However, in MR-VNet, only the second-order nonlinearity is captured, while the first-order nonlinearity is overlooked. This results in reduced attention to local features and a higher risk of gradient explosion. In this paper, we design a VBlock that addresses these issues and approximates nonlinearity without the computational burden of stacking multiple nonlinear functions.

\subsection{Mixture of Experts}
Mixture of Experts (MoE)~\cite{moe6797059} consists of multiple experts and a routing network that combines their outputs using a weighted strategy. Its main goal is to enhance model performance by scaling up parameters while maintaining computational efficiency~\cite{moeYu2024BoostingCL,moe10.5555,moemustafa2022multimodal,moeruiz2021scaling,Lifelong-MoE,moehyChen_Pan_Lu_Fan_Li_2023,DRSformer,moedanYe2022TowardsES}.
HCTFFN~\cite{moehyChen_Pan_Lu_Fan_Li_2023} leverages MoE to highlight features related to spatially varying rain distribution. DRSformer~\cite{DRSformer} incorporates an MoE feature compensator for a unified exploration of joint data and content sparsity. DAN-Net~\cite{moedanYe2022TowardsES} employs MoE to extract degradation information from each input image.  Lifelong-MoE~\cite{Lifelong-MoE} uses pre-trained experts and gates to retain prior knowledge. 
In contrast to the methods mentioned above, we incorporate prior knowledge into DDRE from a well-trained model to estimate spatially variable blur information. This allows the router to adaptively allocate weights to different experts, enabling them to address the varying degradation degrees presented by different blur regions.
\begin{figure*}[htb] % use float package if you want it here
	\centering
	\includegraphics[width=1\textwidth]{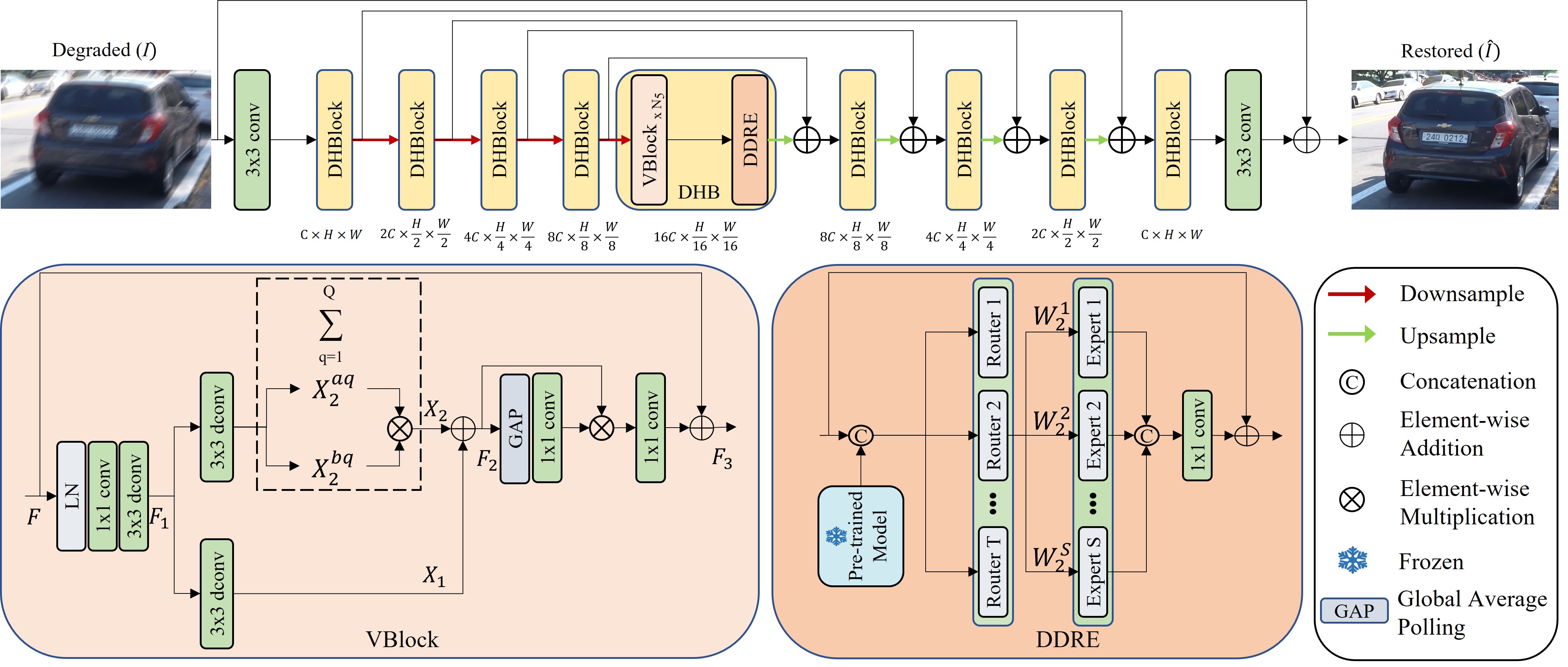}
	\caption{The overall architecture of the proposed differential handling network (DHNet)  for image deblurring mainly consists of the differential handling block (DHBlock), which includes the Volterra block (VBlock) and the degradation degree recognition expert module (DDRE). DDRE is shown in the one-router case for clarity.}
	\label{fig:network}
\end{figure*}

\section{Method}
In this section, we begin with an overview of the entire DHNet pipeline. Following that, we delve into the details of the proposed differential handling block (DHBlock), which serves as the fundamental building unit of our method. This block primarily consists of two key components: the Volterra block (VBlock) and the degradation degree recognition expert module (DDRE).

\subsection{Overall Pipeline} 
Our proposed DHNet, depicted in Figure~\ref{fig:network}, employs a hierarchical encoder-decoder architecture to facilitate effective hierarchical representation learning. Each encoder-decoder includes a DHBlock, which consists of $N_{i \in [1,...,9]}$ VBlocks and a degradation degree recognition expert module (DDRE).
Given a degraded image $\mathbf{I} \in \mathbb{R}^{H \times W \times 3}$, DHNet first applies convolution to extract shallow features $\mathbf{F_{s}} \in \mathbb{R}^{H \times W \times C}$ (where $H$, $W$, and $C$ represent the height, width, and number of channels of the feature map, respectively). 
These shallow features pass through a four-scale encoder sub-network, where the resolution is progressively reduced while the number of channels increases. The resulting in-depth features then move to a middle block, and the deepest features are processed by a four-scale decoder, which gradually restores them to their original size. Finally, we apply convolution to the refined features to generate a residual image $\mathbf{X}\in \mathbb R^{H \times W \times 3}$. This residual image is added to the degraded image to produce the restored image:  $\mathbf{\hat{I}} = \mathbf{X} +\mathbf{I}$.

\subsection{Differential Handling Block}
While existing image deblurring methods have demonstrated commendable performance, they often rely on stacking numerous nonlinear activation functions to approximate nonlinear properties, neglecting the varying degradation degrees across different blurred regions. To address this issue, we design the differential handling block (DHBlock) as a feature extractor, enabling differential processing of diverse blur regions and utilizing Volterra kernel to capture complex input-output relationships. Formally, given the input features at the $(l-1)_{th}$ block $X_{l-1}$, the procedures of DHBlock can be defined as:
\begin{equation}
\begin{aligned}
\label{eq:dh1}
    X_l^{'} &= VBlock_{N_i}(...(VBlock_1(X_{l-1}))...)
    \\
    X_l &= DDRE(X_l^{'}) 
\end{aligned}
\end{equation}
where $N_i$ represents the number of VBlocks in the DHBlock, while  $X_l^{'}$ and $X_l$ denote the outputs from the   $N_i$  Volterra blocks (VBlocks) and the degradation degree recognition expert module (DDRE), which are detailed below.

\subsection{Volterra Block}
The strong performance and versatility of deep learning-based image deblurring models~\cite{FSNet,AdaRevD} can be attributed to their nature as universal approximators. They achieve this by stacking numerous nonlinear activation functions, allowing them to fit any nonlinear function. 
In order to reduce the system complexity caused by an excess of  nonlinear activation functions,  we design a Volterra block (VBlock) that utilizes the Volterra kernel to explore non-linearity within the network. Instead of relying on traditional nonlinear activation functions, VBlock employs higher-order convolutions to enhance linear convolution by enabling interactions between image pixels. As shown in Figure~\ref{fig:network}, unlike MR-VNet~\cite{MR-VNet}, which treats the Volterra filter as a complete block, our VBlock incorporates it as part of a block. The VBlock first captures local features using LN, 1x1, and 3x3 convolutions before entering the Volterra filter, and then applies two 3x3 convolutions for each order branch.  Specifically, our VBlock takes an input tensor $F$, first applies layer normalization (LN), and then encodes channel-wise context by utilizing 1×1 convolutions followed by 3×3 depth-wise convolutions to obtain the feature $F_1$ as follows:
\begin{equation}
\label{vb1}
    F_1 = W_d^0W_p^0LN(F)
\end{equation}
where $W_p^{(\cdot)}$ denotes the $1 \times 1$ point-wise convolution, and $W_d^{(\cdot)}$ represents the $3 \times 3$ depth-wise convolution. 

The resulting feature $F_1$ is then processed by the second-order Volterra kernel to capture complex input-output relationships through interactions between image pixels. In this paper, we compute the second-order Volterra kernel as a product of traditional correlation operations. This representation can achieve an approximation error with arbitrary precision. Moreover, the separability assumption of the kernels enables efficient computation, which is especially advantageous in network settings.
The process is as follows:
\begin{equation}
\begin{aligned}
\label{vb2}
    F_2 &=  X_1 \oplus X_2
    \\
    &= W_d^1 F_1 \oplus (\sum_{q=1}^Q X_2^{aq} \otimes X_2^{bq}) 
    \\
    &= W_d^1 F_1 \oplus (\sum_{q=1}^Q W_d^{2aq} F_1 \otimes W_d^{2bq} F_1)
\end{aligned}
\end{equation}
where $W_d^{2aq}$ and $W_d^{2bq}$ are learnable weight matrices, $Q$ represents the  desired rank of approximation.

\textbf{Theorem 1.}  Any continuous function can be approximated using a Volterra kernel. 

\textbf{Proof.}  Any continuous nonlinear function can be approximated by a polynomial. Similarly to VNNs~\cite{roheda2024volterra}, using a Taylor expansion at $x_0$, the nonlinear function $\sigma(\cdot)$ can be represented as:
\begin{equation}
\begin{aligned}
\label{tp1_1}
    \sigma_t(x) &= f(x_0) + ... + \frac{f^{n}(x_0)}{n!}(x-x_0)^n + R_n(x)
    \\ 
     &= \alpha_0 + \alpha_1x + ...+ \alpha_n x^n + R_n(x)
\end{aligned}
\end{equation}

For a memoryless higher-order Volterra kernel, it can be formulated as:
\begin{equation}
\label{tp1_2}
    \sigma_v(x)  = h_0 + h_1x + ...+ h_n x^n
\end{equation}
where $h_n$ represents the $n_{th}$ order weight, which is learned during the training process. The function $\sigma_v(x)$ can be considered an $n_{th}$ order approximation of $\sigma_t(x)$, with the error defined as:
\begin{equation}
\label{tp1_3}
    error = (\alpha_0 - h_0) + ...+  (\alpha_n - h_n)x^n + R_n(x)
\end{equation}

For a finite polynomial, the absolute error of the expansion can be quantitatively expressed using the Taylor Remainder as follows:
\begin{equation}
\label{tp1_4}
    |error| \leq R_n(x) =  \frac{f^{n+1}(\xi)}{(n+1)!} (x-x_0)^{n+1}
\end{equation}
where $\xi \in (x_0, x)$. Due to the complexity of higher-order Volterra kernels, this paper utilizes a cascade of second-order Volterra kernels to approximate the higher-order behavior (refer to supplementary material for the proof).

\textbf{Theorem 2.} A Volterra kernel provide more adaptive representation than that possible by activation functions.

\textbf{Proof.} For nonlinear activation functions such as ReLU, sigmoid, and tanh, the coefficients $\alpha_n$ in their Taylor expansions are predetermined\footnote{The term "predetermined" refers to a fixed value that remains constant.}. Specifically, a sigmoid activation can be approximated as:
\begin{equation}
\label{tp2_1}
    \sigma_s(x)  = \frac{1}{1+e^{-x}} = \frac{1}{2} + \frac{1}{4}x - \frac{1}{48}x^3 +...
\end{equation}
The expansion coefficient $h_n$ of the Volterra kernel can be continuously adjusted and learned during training, allowing for greater adaptability and more precise mapping of complex scenes. As a result, the feature $F_2$ in Eq.~\ref{vb2} exhibits nonlinear characteristics and offers greater flexibility compared to features obtained through  nonlinear activation functions (e.g. sigmoid). Finally, we apply channel attention to $F_2$ to obtain the final output features $F_3$ of VBlock as follows:
\begin{equation}
\label{vb3}
    F_3 = W_p^2(F_2 \otimes (W_p^1GAP(F_2))) \oplus F
\end{equation}

\subsection{Degradation Degree Recognition Expert Module}
While existing image deblurring methods~\cite{MR-VNet} have demonstrated strong performance, they overlook the inconsistency in the degradation degree across different regions. As illustrated in Figure~\ref{fig:ques}, the degree of degradation in a blurred image can differ significantly between areas.  Thus, it is clearly unreasonable for these methods to assume that all degraded areas share the same degree of degradation. 
In contrast to AdaRevD~\cite{AdaRevD}, which relies on a limited set of predefined categories and a fixed blur patch size, DDRE first incorporates prior knowledge from a well-trained model to estimate the spatially variable blur information. This enables DDRE to allow the router to map the learned degradation representation and assign weights to experts according to both the degree of degradation and the size of the regions.  

Different from the conventional mixture of experts~\cite{moeYu2024BoostingCL, Lifelong-MoE} that select only one router and a subset of experts for each router, our DDRE utilizes all routing paths, dynamically assigning weights to each expert along every router. Each expert comprises convolutions with varying receptive field sizes, allowing for the identification of degraded regions of different sizes. For simplicity, Figure~\ref{fig:network} illustrates a scenario with just one router. Given an input feature map $F_4$, we first integrate prior knowledge from a well-trained model and keeps it frozen during training for lower memory consumption as:
\begin{equation}
\label{ddre1}
    F_5 = W_p^3[F_4, P]
\end{equation}
where $P$ is the prior knowledge and $[\cdot]$ denotes the concatenation. The feature map $F_5$ that aggregate the degradation information is then input to each router, which assigns weights to each expert as follows:
\begin{equation}
\label{ddre2}
    F_6 = [W_j^k E_k(F_5)] 
\end{equation}
where $W_j^k$  represents the learnable weight coefficient assigned by each router $j \in \{1,2,...,T\}$ to each expert $k \in \{1,2,...,S\}$, and $E$ is the expert operations. Finally, the output of the DDRE is calculated by:
\begin{equation}
\label{ddre3}
    F_7 = W_p^4F_6 \oplus F_4 
\end{equation}

With this design, DDRE can adaptively handle varying degradation degrees across different sizes of degraded regions.
\section{Experiments}

In this section, we outline the experimental settings and present both qualitative and quantitative comparisons between DHNet and other state-of-the-art methods. We then conduct ablation studies to highlight the effectiveness of our approach. The best and second-best results in the tables are indicated in \textbf{bold} and \underline{underlined} formats, respectively.

\subsection{Experimental Settings}

\subsubsection{\textbf{Training details}}
We train separate models for different tasks, and unless otherwise specified, the following parameters are utilized.  In our model, $N_{i \in [1,2,3,4,5,6,7,8,9]}$ are set to $\{1,1,1,28,1,1,1,1,1\}$. In terms of VBlock, we set $Q = 4$ (Eq.~\ref{vb2}).
In terms of DDRE, we set $ S = 5$ for the number of experts and $T = 4$ for the number of routers. 
We use the Adam optimizer~\cite{2014Adam} with parameters $\beta_1=0.9$ and $\beta_2=0.999$. The initial learning rate is set to $5 \times 10^{-4}$ and gradually reduced to $1 \times 10^{-6}$ using the cosine annealing strategy~\cite{2016SGDR}. The batch size is chosen as $32$, and patches of size $256 \times 256$ are extracted from training images. Data augmentation includes both horizontal and vertical flips. Given the high complexity of image motion deblurring, we set the number of channels to 32 for DHNet and 64 for DHNet-B. 
In DHNet, we do not use a pre-trained model in the DDRE, whereas DHNet-B utilizes UFPNet~\cite{UFPNetFang_2023_CVPR}.

\begin{figure*} % use float package if you want it here
	\centering
	\includegraphics[width=1\linewidth]{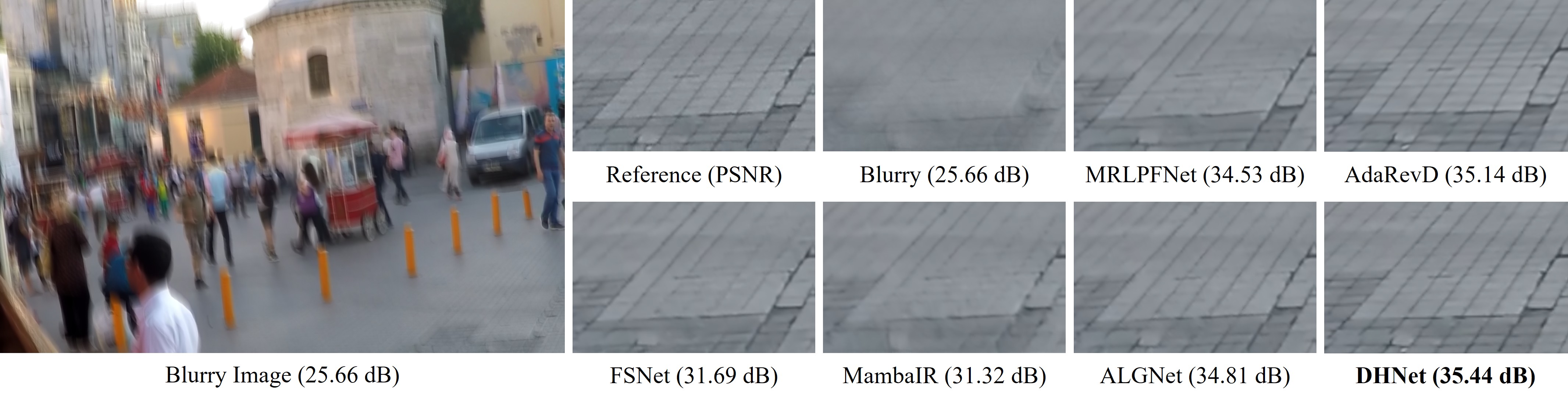}
	\caption{Image motion deblurring comparisons on the GoPro dataset~\cite{Gopro}. Our DHNet recovers image with clearer details. }
	\label{fig:blurm}
\end{figure*}

\subsubsection{Datasets}

We validate the effectiveness of our method using the GoPro dataset~\cite{Gopro}, which consists of 2,103 training image pairs and 1,111 evaluation pairs, in line with recent approaches~\cite{FSNet}. To evaluate the generalizability of our model, we  apply the GoPro-trained model to the HIDE~\cite{HIDE} dataset, containing 2,025 images specifically designed for human-aware motion deblurring. Both the GoPro and HIDE datasets are synthetically generated. Additionally, we assess our method's performance on real-world images using the RealBlur~\cite{realblurrim_2020_ECCV} dataset, which contains 3,758 training image pairs and 980 testing pairs, divided into two subsets: RealBlur-J and RealBlur-R.

\begin{table}
\centering
\caption{Quantitative evaluations of the proposed approach against state-of-the-art motion deblurring methods. Our DHNet and DHNet-B are trained only on the GoPro dataset. \label{tb:deblurgh}}
\resizebox{\linewidth}{!}{
\begin{tabular}{ccccc}
    \hline
    \multicolumn{1}{c}{} & \multicolumn{2}{c}{GoPro}  & \multicolumn{2}{c}{HIDE} 
    \\
   Methods & PSNR $\uparrow$ & SSIM $\uparrow$ & PSNR $\uparrow$ & SSIM $\uparrow$   
    \\
    \hline\hline
    MPRNet~\cite{Zamir2021MPRNet} & 32.66 & 0.959 & 30.96 & 0.939 
    \\
    Restormer~\cite{Zamir2021Restormer} & 32.92 & 0.961 & 31.22 & 0.942 
    \\
     Uformer~\cite{Wang_2022_CVPR} &32.97 & 0.967 &30.83 &\underline{}{0.952} 
     \\
       NAFNet-32~\cite{chen2022simple}&32.83&0.960&-&-
    \\
    NAFNet-64~\cite{chen2022simple}&33.62&0.967&-&-
    \\
    IRNeXt~\cite{IRNeXt} &33.16 &0.962 &- & - 
    \\
    SFNet~\cite{SFNet} &33.27 &0.963 &31.10 & 0.941 
    \\
    DeepRFT+~\cite{fxint2023freqsel} &  33.23 &0.963 &31.66 &0.946
    \\
    UFPNet~\cite{UFPNetFang_2023_CVPR} &34.06 &0.968 &31.74 &0.947
    \\
    MRLPFNet~\cite{MRLPFNet} &34.01 &0.968 &31.63 &0.947
    \\
     MambaIR~\cite{guo2024mambair}&33.21 &0.962 &31.01 &0.939
     \\
     ALGNet-B~\cite{alggao2024learning} &34.05 &0.969 &31.68 &\underline{0.952}
     \\
     MR-VNet~\cite{MR-VNet} & 34.04 & 0.969 & 31.54 & 0.943
     \\
    FSNet~\cite{FSNet} &33.29&0.963 &31.05 & 0.941 
    \\
    AdaRevD-B~\cite{AdaRevD} & 34.50 &0.971 &32.26 &\underline{0.952}
    \\
    AdaRevD-L~\cite{AdaRevD} &\underline{34.60} &\underline{0.972} &\underline{32.35} &\textbf{0.953} 
    \\
    \hline
    \textbf{DHNet(Ours)}& 33.28	&0.964	&31.75	&0.948
    \\
    \textbf{DHNet-B(Ours)} & \textbf{34.75} & \textbf{0.973} & \textbf{32.37} & \textbf{0.953}
    \\
    \hline
\end{tabular}}
\end{table}

\begin{table}
\centering
\caption{ Quantitative evaluations of the proposed approach against
state-of-the-art methods on the real-word dataset RealBlur~\cite{realblurrim_2020_ECCV}. }
\label{tb:0deblurringreal}
\resizebox{\linewidth}{!}{
\begin{tabular}{ccccc}
    \hline
    \multicolumn{1}{c}{} & \multicolumn{2}{c}{RealBlur-R}  & \multicolumn{2}{c}{RealBlur-J} 
    \\
   Methods & PSNR $\uparrow$ & SSIM $\uparrow$ & PSNR $\uparrow$ & SSIM $\uparrow$   
    \\
    \hline\hline
    DeblurGAN-v2~\cite{deganv2} & 36.44 & 0.935& 29.69& 0.870
\\
    MPRNet~\cite{Zamir2021MPRNet} & 39.31 & 0.972 & 31.76 & 0.922
   \\
   DeepRFT+~\cite{fxint2023freqsel}&39.84 &0.972 &32.19 &0.931
\\
Stripformer~\cite{Tsai2022Stripformer} & 39.84 & 0.975 & 32.48 & 0.929
\\
FFTformer~\cite{kong2023efficient}&40.11& 0.973 &32.62 &0.932
\\
UFPNet~\cite{UFPNetFang_2023_CVPR} &40.61 &0.974 &33.35 &0.934 
\\
 MambaIR~\cite{guo2024mambair}& 39.92 & 0.972 & 32.44 & 0.928
 \\
 ALGNet~\cite{alggao2024learning} & 41.16 &0.981 &32.94 &0.946
 \\
 MR-VNet~\cite{MR-VNet} & 40.23 & 0.977 &32.71 & 0.941
 \\
 AdaRevD-B~\cite{AdaRevD} &41.09 &0.978 &33.84 &0.943
 \\
 AdaRevD-L~\cite{AdaRevD} &41.19 &0.979 &\underline{33.96} &0.944
 \\
 \hline
    \textbf{DHNet(Ours)}& \underline{41.30} & \textbf{0.984} & 33.78 & \underline{0.952}
    \\
     \textbf{DHNet-B(Ours)}& \textbf{41.33} & \underline{0.983} & \textbf{34.28} & \textbf{0.953}
    \\
    \hline
\end{tabular}}
\end{table}

\subsection{Experimental Results}
\subsubsection{ \textbf{Evaluations on the synthetic dataset.}}
We present the performance of various image deblurring methods on the synthetic GoPro~\cite{Gopro} and HIDE~\cite{HIDE} datasets in Table~\ref{tb:deblurgh}. Overall, our DHNet outperforms competing approaches, yielding higher-quality images with superior PSNR and SSIM values. Specifically, compared to our baseline model, NAFNet~\cite{chen2022simple}, DHNet improves performance by 0.45 dB and 1.13 dB with 32 and 64 channels, respectively. In comparison to the previous best method, AdaRevD-L~\cite{AdaRevD}, our DHNet-B shows an improvement of 0.15 dB on the GoPro~\cite{Gopro} dataset. Notably, although our model was trained exclusively on the GoPro~\cite{Gopro} dataset, it still achieves state-of-the-art results (32.37 dB in PSNR) on the HIDE~\cite{HIDE} dataset, demonstrating its excellent generalization capability.
Furthermore, Figure~\ref{fig:param} shows that DHNet not only achieves state-of-the-art performance but also reduces computational costs. The performance continues to improve with an increase in model size, highlighting the scalability of our approach. Finally, Figure~\ref{fig:blurm} presents deblurred images from the evaluation methods, where our model produces more visually pleasing results.

\begin{figure*} % use float package if you want it here
	\centering
	\includegraphics[width=1\linewidth]{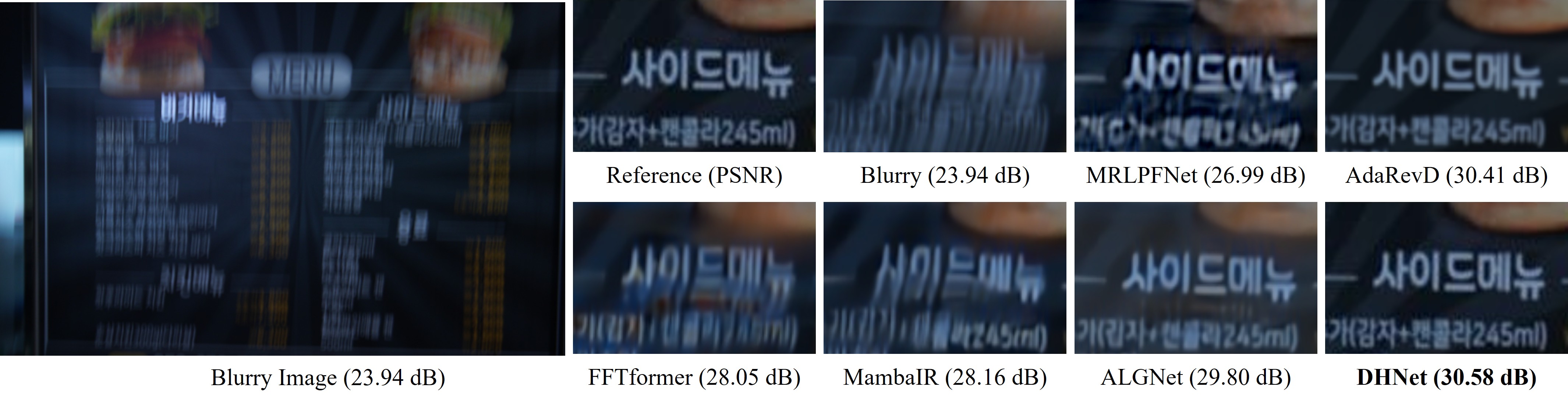}
	\caption{Image motion deblurring comparisons on the RealBlur dataset~\cite{realblurrim_2020_ECCV}. Our DHNet recovers perceptually faithful images. }
	\label{fig:realm}
\end{figure*}

\subsubsection{ \textbf{Evaluations on the real-world dataset.}}
In addition to evaluating on synthetic datasets, we further assess the performance of our DHNet on real-world images from the RealBlur dataset~\cite{realblurrim_2020_ECCV}. As shown in Table~\ref{tb:0deblurringreal}, our method produces deblurred images with superior PSNR and SSIM scores. Specifically, compared to the previous best method, AdaRevD-L~\cite{AdaRevD}, our approach achieves improvements of 0.14 dB and 0.32 dB on the RealBlur-R and RealBlur-J datasets, respectively. Figure~\ref{fig:realm} demonstrates that DHNet produces clearer images with finer details and structures, outperforming the competing methods.

\begin{table}
    \centering
       \caption{Ablation study on individual components of the
proposed DHNet.}
    \label{tab:abl1}
    
    \begin{tabular}{ccccc}
    \hline
         Net&VBlock& DDRE  & PSNR & $\triangle$ PSNR
         \\
         \hline
         (a)& &  &    33.62 & -
         \\
         (b)& \ding{52}&     & 34.05 & +0.43
         \\
         (c)&  &  \ding{52}  &  34.32 & +0.70
         \\
         (d)& \ding{52}&    \ding{52}&  34.75  & +1.13
         \\
         \hline
    \end{tabular}
\end{table}

\subsection{Ablation Studies}
In this section, we first demonstrate the effectiveness of the proposed modules and then investigate the effects of different designs for each module. 

\subsubsection{Effects of Individual Components} 

To evaluate the effectiveness of each module, we use NAFNet~\cite{chen2022simple} as our baseline model and subsequently replace or add the modules we have designed. As shown in Table~\ref{tab:abl1}(a), the baseline achieves  33.62 dB PSNR. Each module combination leads to a corresponding performance improvement. Specifically, replacing NAFBlock~\cite{chen2022simple} with our VBlock enhances the performance by 0.43 dB (Table~\ref{tab:abl1}(b)). Adding the DDRE module to the original NAFNet~\cite{chen2022simple} can significantly boost the model's performance from 33.62 dB to 34.32 dB (Table~\ref{tab:abl1}(c)). When all modules are combined (Table~\ref{tab:abl1}(d)), our model achieves a 1.13 dB improvement over the original baseline.

\begin{figure*}[htb] % use float package if you want it here
	\centering
	\includegraphics[width=1\linewidth]{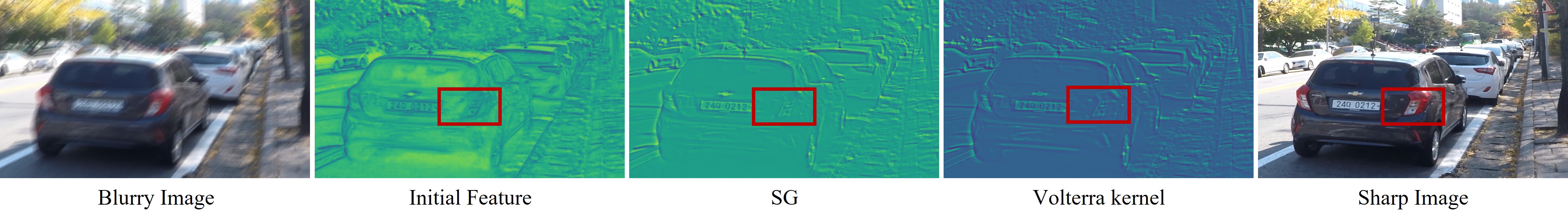}
	\caption{Effect of the proposed VBlock. Zoom in for the best view.}
	\label{fig:vbabl}
\end{figure*}

\begin{table}
    \centering
       \caption{Results of alternatives to VBlock.}
    \label{tab:vbl1}
    \resizebox{\linewidth}{!}{
    \begin{tabular}{ccccccc}
    \hline
         Net& SG~\cite{chen2022simple} &Volterra& Q  & PSNR & $\triangle$ PSNR &MACs(G)
         \\
         \hline
         (a)& \ding{52} &  &0  &34.12 & - & 106
         \\
         (b)& & \ding{52} &1    &34.25 & +0.13 &107
         \\
         (c)&  &  \ding{52} & 2  & 34.42 & +0.30 &109
         \\
         (d)& &    \ding{52}& 4 & 34.75 & +0.63 & 111
         \\
         (e)& &    \ding{52}& 8 &34.76 & +0.64  & 119
         \\
         \hline
    \end{tabular}}
\end{table}

\begin{figure*}[htb] % use float package if you want it here
	\centering
	\includegraphics[width=1\linewidth]{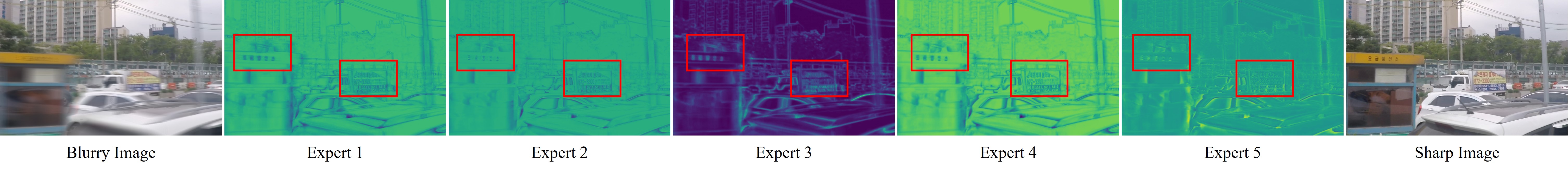}
	\caption{Effect of the proposed DDRE. Zoom in for the best view.}
	\label{fig:ddreabl}
\end{figure*}
\subsubsection{Design Choices for VBlock} 
VBlock facilitates the generation of nonlinear interactions through the interactions between image pixels. To evaluate the effectiveness of VBlock, we first use SG~\cite{chen2022simple} to replace the Volterra kernel and then examine the impact of varying the kernel rank. As shown in Table~\ref{tab:vbl1}, the SG~\cite{chen2022simple} achieves a PSNR of 34.12 dB, and performance improves when we implement our Volterra kernel. We further visualize the feature maps in Figure~\ref{fig:vbabl} to highlight the advantages of our Volterra kernel compared to SG~\cite{chen2022simple}. The results clearly demonstrate that features obtained with the Volterra kernel are more detailed (as indicated by the red box) and better suited for handling complex scenes. Additionally, we analyze the effect of the $Q$ value on the performance. As we increase the rank $Q$, performance consistently improves, but this also increases system complexity. To achieve a balance between efficiency and performance, we choose an experimental setting with $Q = 4$.

\subsubsection{Design Choices for DDRE} 

To evaluate the effectiveness of DDRE, we conduct experiments with various model variants, as shown in Table~\ref{tab:ddrel1}. When we introduce experts (Table~\ref{tab:ddrel1} (b)) into the baseline model (Table~\ref{tab:ddrel1} (a)), performance improves from 34.05 to 34.26, highlighting the model's capability to address degradation in a differentiated manner. Further enhancement occurs when we incorporate external degradation pattern knowledge, leading to even greater performance gains (Table~\ref{tab:ddrel1} (c) and (d)). Additionally, we observe that different pre-trained models (NAFNet~\cite{chen2022simple}, UFPNet~\cite{UFPNetFang_2023_CVPR}) yield varying performance outcomes. We also investigate the impact of the number of routers and experts on the experimental results. Our findings indicate that the combination of $ S = 5$ and $ T = 4$ (Table~\ref{tab:ddrel1} (d)) yields the best results. A smaller number (Table~\ref{tab:ddrel1} (e) (g)) fails to adequately identify the degradation degree. While an excessive number (Table~\ref{tab:ddrel1} (f) (h)) tends to focus on the severely degraded regions, and the lightly degraded regions will receive relatively less attention, ultimately hindering their restoration. 
To further demonstrate the effectiveness of our DDRE module, we visualize the features of a routing path weighted over all experts, as shown in Figure~\ref{fig:ddreabl}. It is clear that each expert handles the inconsistent degradation region as well as  the degree of restored degradation.

\begin{table}
    \centering
       \caption{Results of alternatives to DDRE, where $S$ and $T$ denote the number of experts and routers, respectively.}
    \label{tab:ddrel1}
    \resizebox{\linewidth}{!}{
    \begin{tabular}{ccccccc}
    \hline
         \multirow{2}{*}{Net}&\multicolumn{2}{c}{Pre-trained}& \multirow{2}{*}{T}  & \multirow{2}{*}{S}& \multirow{2}{*}{PSNR} & \multirow{2}{*}{$\triangle$ PSNR}
         \\
         & NAFNet~\cite{chen2022simple}&UFPNet~\cite{UFPNetFang_2023_CVPR} & & & & 
         \\
         \hline
         (a)& &   &0  &0  & 34.05 & -
         \\
         (b)& &  &4  &5 &34.26 & +0.21
         \\
         (c)&\ding{52}  &  &4  & 5  & 34.47 & +0.35
         \\
         (d)& &  \ding{52}& 4 & 5 &34.75 & +0.70
         \\
         (e)& &  \ding{52}&4 &3 &34.56 & +0.51
         \\
         (f)& &  \ding{52}&4 &8 &34.61 & +0.56
          \\
         (g)& &  \ding{52}&2 &5 &34.51 & +0.46
         \\
          (h)& &  \ding{52}&6 &5 &34.57 & +0.52
         \\
         \hline
    \end{tabular}}
\end{table}

% 证明确实每个专家关注的区域不同    
\section{Conclusion}
In this paper, we propose a differential handling network (DHNet) aimed at accurately deblurring images by processing different blur regions. Specifically, we design a DHBlock consisting of VBlock and DDRE. The VBlock utilizes the Volterra kernel  to explore non-linearity within the network to map complex input-output relationships. Meanwhile, the DDRE integrates prior knowledge from a well-trained model to estimate spatially variable blur information, allowing the router to map the learned degradation representation and assign weights to experts based on the degree of degradation and the size of the regions. Experimental results demonstrate that our DHNet outperforms state-of-the-art approaches.
{
    \small
    \bibliographystyle{ieeenat_fullname}
    \bibliography{main}
}
\clearpage
\setcounter{page}{1}
\maketitlesupplementary

\section{Overview}
The supplementary material is composed of:

Motivation Analysis: Sec.\ref{sec:ma}

More Proof for VBlock: Sec.\ref{sec:mpvb}

Loss Function: Sec.\ref{sec:loss}

More Ablation Studies: Sec.\ref{sec:mas}

Additional Visual Results: Sec.\ref{sec:Visual}

\section{Motivation Analysis}
\label{sec:ma}
\begin{figure}[htb] % use float package if you want it here
	\centering
	\includegraphics[width=1\linewidth]{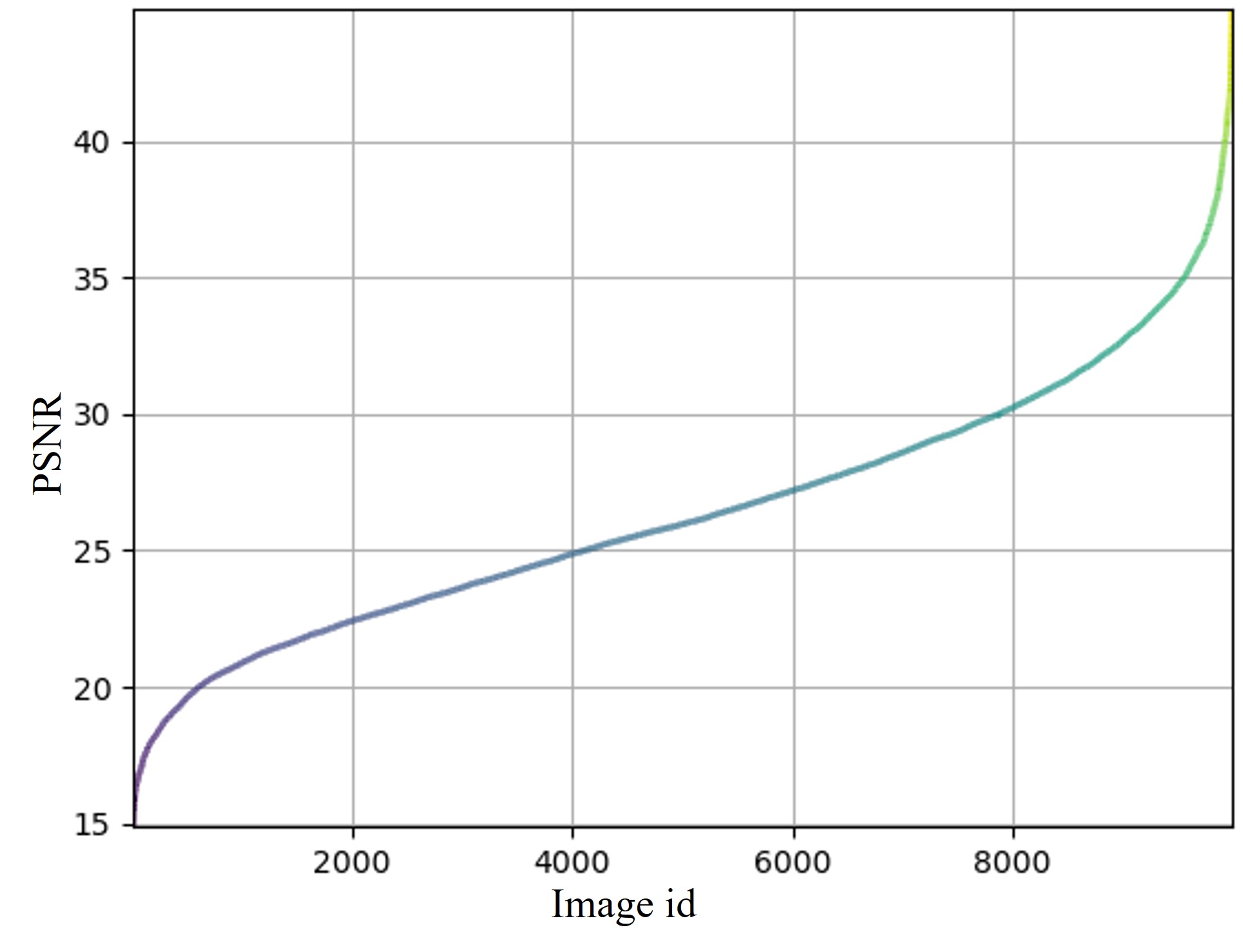}
	\caption{The ranked PSNR curve of the different blur region from GoPro~\cite{Gopro} test set.}
	\label{fig:diffr}
\end{figure}

\begin{figure}[htb] % use float package if you want it here
	\centering
	\includegraphics[width=1\linewidth]{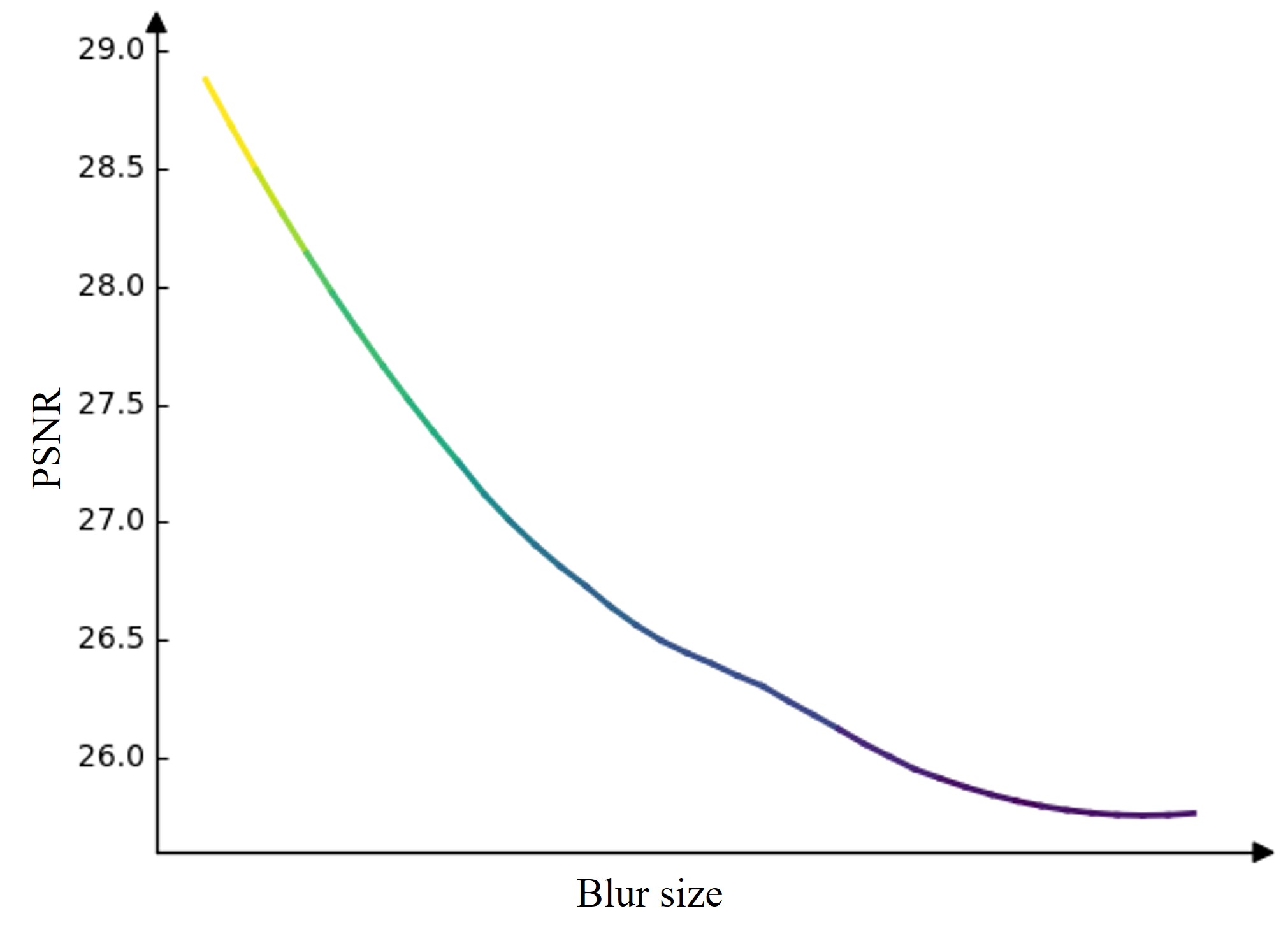}
	\caption{The PSNR curve of the different blur size from GoPro~\cite{Gopro} test set.}
	\label{fig:diffs}
\end{figure}

Although existing image deblurring methods~\cite{kong2023efficient, MR-VNet} have demonstrated strong performance, they fail to account for the varying degrees of degradation across different regions. As shown in Figure~\ref{fig:diffr}, the blur intensity differs across regions of the image. Additionally, Figure~\ref{fig:diffs} highlights that larger blurred regions are more challenging to recover. To address the first issue, AdaRevD~\cite{AdaRevD} introduces a classifier to assess the degradation degree of image patches. However, AdaRevD~\cite{AdaRevD} relies on a limited set of predefined categories and a fixed blur patch size, which restricts its ability to effectively adapt to different degradation degrees across varying patch sizes. This limitation prevents it from adequately solving the second problem. 

\begin{figure*}[htb] % use float package if you want it here
	\centering
	\includegraphics[width=1\linewidth]{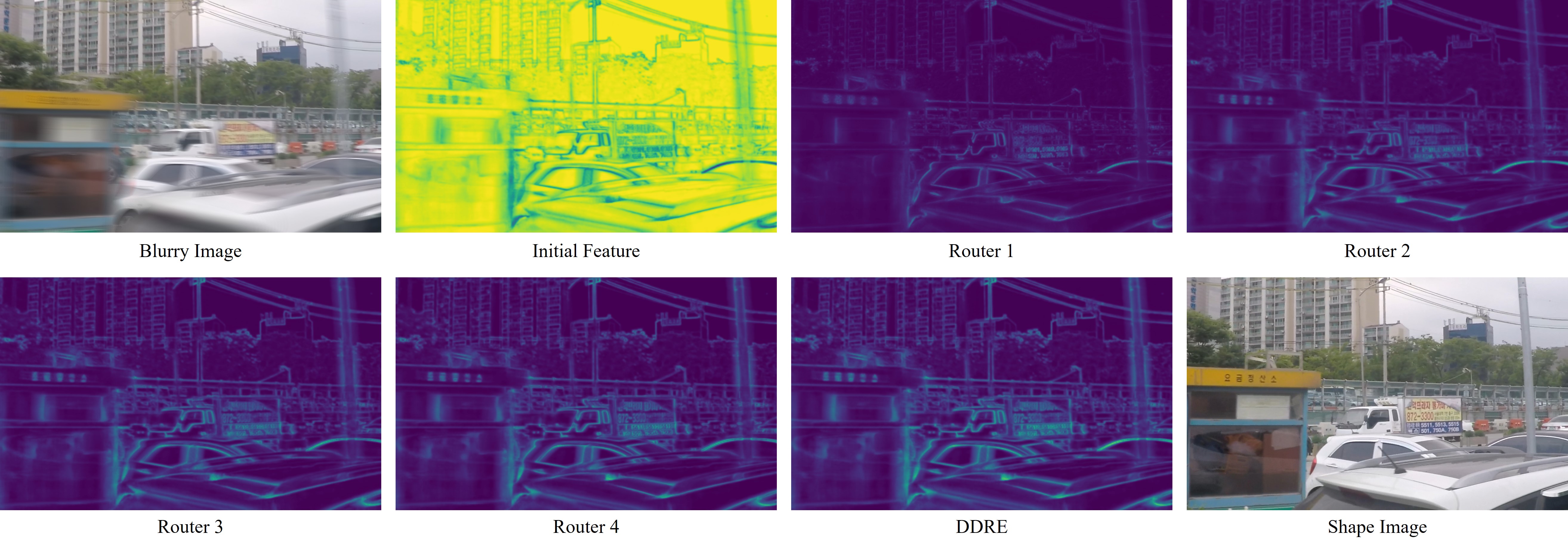}
	\caption{The internal features of DDRE.}
	\label{fig:ddrerr}
\end{figure*}

In this paper, we propose the degradation degree recognition expert (DDRE) module which enables the model to adaptively handle varying degrees of degradation in blurred regions. Unlike conventional mixture of experts methods~\cite{moeYu2024BoostingCL, Lifelong-MoE}, where each router selects only one expert and a subset of experts for processing, our DDRE first integrates prior knowledge from a well-trained model to estimate spatially varying blur information. It then utilizes all available routing paths, dynamically assigning weights to each expert along every path. As shown in Table~\ref{tab:ddreexp}, each expert is configured with convolutions of varying receptive field sizes, allowing it to identify degraded regions at different scales. To further validate the effectiveness of the DDRE module, we visualize the feature map in Figure~\ref{fig:ddrerr}. Compared to the initial feature map, the DDRE module recovers images with clearer details, such as the phone number on the white car advertisement. Moreover, the different routers exhibit varying abilities to handle blur degradation, demonstrating that our method can adaptively address blur with different degrees of degradation.

\begin{table}
    \centering
       \caption{The configuration of each expert.}
    \label{tab:ddreexp}

    \begin{tabular}{cc}
    \hline
        Expert &  Configuration
         \\
         \hline
         (1)& 1x1 depthwise separable convolution
         \\
         (2)& 3x3 depthwise separable convolution
         \\
         (3)&5x5 depthwise separable convolution
         \\
         (4)& 7x7 depthwise separable convolution
         \\
         (5)& 9x9 depthwise separable convolution
         \\
         \hline
    \end{tabular}
\end{table}

\section{More Proof for VBlock}
\label{sec:mpvb}
Our theoretical proof ideas are borrowed from VNNs~\cite{roheda2024volterra}. A Volterra kernel with $ L $  terms, it can be expressed as:
\begin{equation}
\begin{aligned}
\label{sp1_1}
   y_k  = h_0 + \sum_{d=1}^n \sum_{r_1 = 0}^{L-1} ... \sum_{r_d = 0}^{L-1} h_d(r_1...r_d) \prod_{j=1}^dx(k-r_j)
\end{aligned}
\end{equation}
where $d$ represents the order, $n$ is  the maximum number of $b$, and $r_d$ denotes the memory delay.

For image data, the feature value at location $[x_1, x_2]$ in feature map $F$ is computed using a 2D version of the Volterra kernel, as expressed:
\begin{equation}
\begin{aligned}
\label{sp1_2}
 & F_z \left[ \begin{array}{c}
       x_1
        \\
        x_2
  \end{array}\right ] 
  = V_z(F_{z-1} \left[ \begin{array}{c}
       x_1 - p_1 : x_1 + p_1
        \\
        x_2 - p_2 : x_2 + p_2
  \end{array}\right ] )
  \\
  &= \sum_{r_{11}, r_{21}}h_1 \left[ \begin{array}{c}
       r_{11}
        \\
        r_{21}
  \end{array}\right ]x\left[ \begin{array}{c}
       x_1 - r_{11}
        \\
        x_2 - r_{21}
  \end{array}\right ]
  \\
  &+
  \sum_{\substack{ r_{11}, r_{21} \\ r_{12}, r_{22}}}h_2 \left[ \begin{array}{c}
       r_{11}
        \\
        r_{21}
  \end{array}\right ]\left[ \begin{array}{c}
       r_{12}
        \\
        r_{22}
  \end{array}\right ]
  x\left[ \begin{array}{c}
       x_1 - r_{11}
        \\
        x_2 - r_{21}
  \end{array}\right ]
   x\left[ \begin{array}{c}
       x_1 - r_{12}
        \\
        x_2 - r_{22}
  \end{array}\right ] 
  \\
  &+...
\end{aligned}
\end{equation}
where $V_z$ denotes the $z_{th}$ layer of the Volterra  kernel, $r_{1i} \in [-p_1,p_1]$ and $r_{2i} \in [-p_2,p_2]$ represent spatial translations in the horizontal and vertical directions, respectively. The  complexity of a $n_{-th}$ order Volterra kernel is computed as:
\begin{equation}
\begin{aligned}
\label{sp1_3}
  \sum_{d=1}^n ( L[2p_{1}+1][2p_{2}+1])^d
\end{aligned}
\end{equation}

The Volterra kernel described earlier is significantly more expressive because it captures higher-order relationships among inputs. However, its computation requires iterated integrals and does not have an efficient GPU implementation. To overcome this limitation, we approximate higher-order behavior by utilizing a convex combination of the first-order and second-order terms of the Volterra kernel.

\textbf{Theorem 3.} A cascade of second-order Volterra kernels can approximate the higher-order behavior.

\textbf{Proof.} The second-order Volterra kernel can be formulated as:
\begin{equation}
\label{sp1_4}
    \sigma_{v2}(x)  = h_0 + h_1x + h_2 x^2
\end{equation}

After feeding the output of the second-order Volterra kernel $\sigma_{v2}(\cdot)$  back into another second-order Volterra kernel, we obtain an output that reaches up to the fourth order:
\begin{equation}
\begin{aligned}
\label{sp1_4}
    \sigma_{v2}(\sigma_{v2}(x))  &= \sigma_{v2}(h_0 + h_1x + h_2 x^2)
    \\
    & = h_0^{'} + h_1^{'}x + h_2^{'} x^2 + h_3^{'}x^3 + h_4^{'}x^4
\end{aligned}
\end{equation}
where:
\begin{equation}
\begin{aligned}
\label{sp1_5}
   h_0^{'}  &=  h_0 + h_0^2h_2
   \\
   h_1^{'}& = h_0h_1 +2h_0h_1h_2
   \\
   h_2^{'} &=  h_1^{2}h_2 + h_1^2
   \\
   h_3^{'} &= 2 h_1h_2h_2^{2} + h_1h_2
   \\
    h_4^{'} &= h_2^3
\end{aligned}
\end{equation} 

Therefore, we can utilize $K$ instances of the second-order Volterra kernel to implement an $n_{-th}$ order Volterra kernel, where $n = 2^{2^{K-1}}$.

\textbf{Theorem 4.} The complexity of an $n_{-th}$ order Volterra kernel, implemented using cascaded second-order Volterra kernels, is calculated as follows:

\begin{equation}
\begin{aligned}
\label{sp1_6}
  \sum_{k=1}^K[(L_k[2p_{1k}+1][2p_{2k}+1]) + ( L_k[2p_{1k}+1][2p_{2k}+1])^2]
\end{aligned}
\end{equation}

\begin{figure*} % use float package if you want it here
	\centering
	\includegraphics[width=1\linewidth]{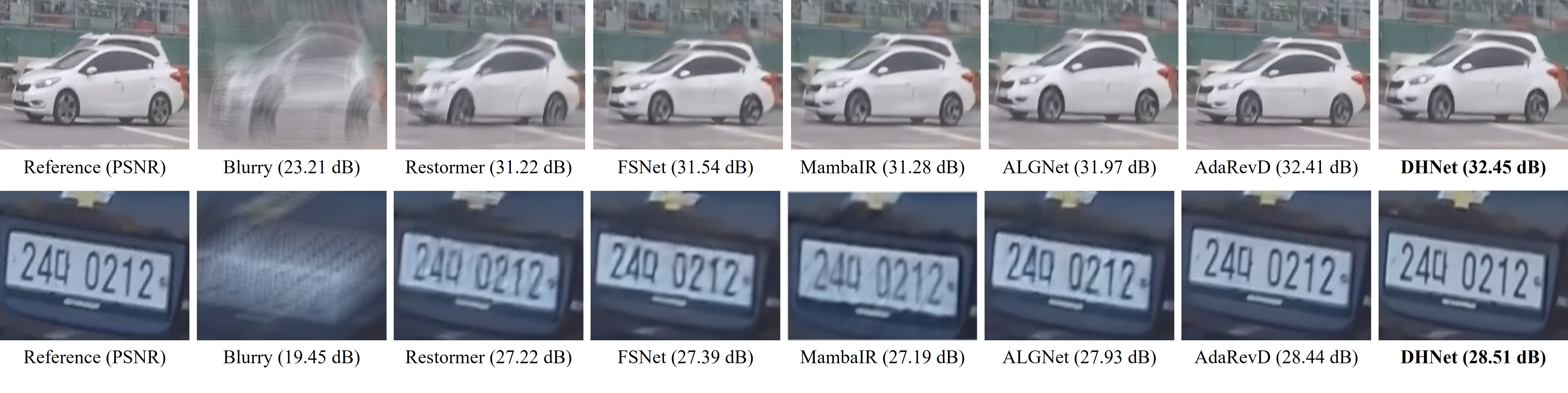}
	\caption{Comparison of image motion deblurring on the GoPro dataset~\cite{Gopro}.}
	\label{fig:mgopr}
\end{figure*}

\begin{figure*} % use float package if you want it here
	\centering
	\includegraphics[width=1\linewidth]{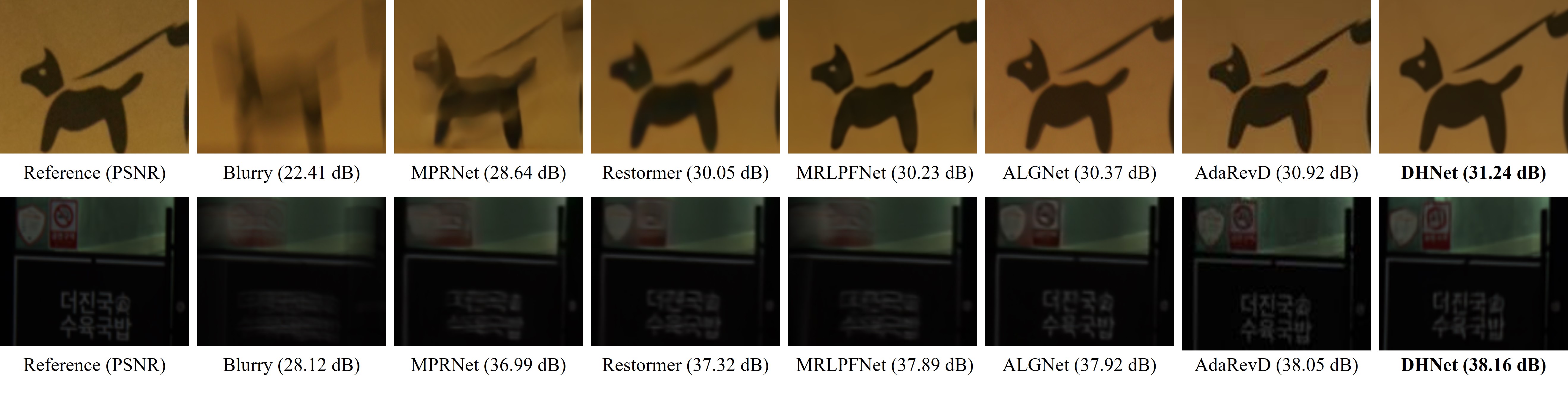}
	\caption{Comparison of image motion deblurring on the RealBlur dataset~\cite{realblurrim_2020_ECCV}.}
	\label{fig:mreal}
\end{figure*}

\begin{figure*} % use float package if you want it here
	\centering
	\includegraphics[width=1\linewidth]{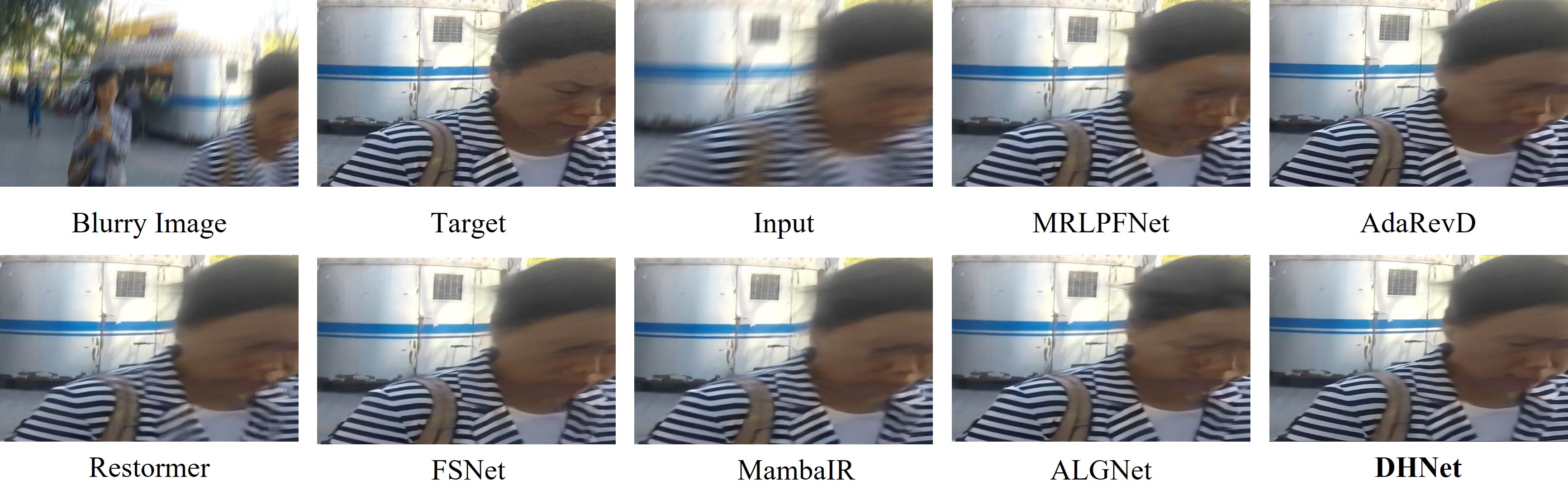}
	\caption{Comparison of image motion deblurring on the HIDE dataset~\cite{HIDE} among MRLPFNet~\cite{MRLPFNet}, AdaRevD~\cite{AdaRevD}, Restormer~\cite{Zamir2021Restormer}, FSNet~\cite{FSNet}, MambaIR~\cite{guo2024mambair}, ALGNet~\cite{alggao2024learning}, and our DHNet.}
	\label{fig:mhide}
\end{figure*}

\textbf{Proof.}  It follows from Eq.~\ref{sp1_3} that, for a second-order Volterra kernel, the number of parameters required is
$[(L_k[2p_{1}+1][2p_{2}+1]) + ( L_k[2p_{1}+1][2p_{2}+1])^2]$. When we utilize $K$ times of the second-order Volterra kernel to implement an $n_{-th}$ order Volterra kernel, it will lead to Eq.~\ref{sp1_6}, which is significantly lower than Eq.~\ref{sp1_3}.

The first order Volterra kernel is similar to the convolutional layer in the conventional CNNs. The second order kernel may be approximated as the concept of separable kernels as :
\begin{equation}
\label{sp1_7}
    W_d^{2}  = \sum_{q=1}^Q W_d^{2aq}  \otimes W_d^{2bq} 
\end{equation}
where $Q$ represents the  desired rank of approximation, $W_d^{2} \in \mathbb{R}^{(2p_1+1)\times (2p_2+1) \times (2p_1+1) \times (2p_2+1)}, W_d^{2aq}\in \mathbb{R}^{(2p_1+1)\times (2p_2+1) \times 1}$ and $W_d^{2bq} \in \mathbb{R}^{1 \times (2p_1+1)\times (2p_2+1)}$. A larger $Q$ will provide a better approximation of the second order kernel. This is easier to implement with a convolutional library in the first place. Secondly, the complexity is reduced from  Eq.~\ref{sp1_6} to:
\begin{equation}
\begin{aligned}
\label{sp1_8}
  \sum_{k=1}^K[(L_k[2p_{1k}+1][2p_{2k}+1]) + 2Q( L_k[2p_{1k}+1][2p_{2k}+1])]
\end{aligned}
\end{equation}

Therefore, we can adjust the value of $Q$ to strike a balance between performance and acceptable computational complexity. We also illustrate the impact of different  $Q$ values on the model in the ablation experiments presented in the main text.

\section{Loss Function}
\label{sec:loss}
To optimize the proposed network DHNet by minimizing the following loss function: 
\begin{equation}
\begin{aligned}
\label{eq:loss1}
L &= L_{c}(\hat{I},\overline I)  + \delta L_{e}(\hat{I},\overline I) + \lambda L_{f}(\hat{I},\overline I))
\\
L_{c} &= \sqrt{||\hat{I} -\overline I||^2 + \epsilon^2}
\\
L_{e} &= \sqrt{||\triangle \hat{I} - \triangle \overline I||^2 + \epsilon^2}
\\
L_{f} &= ||\mathcal{F}(\hat{I})-\mathcal{F}(\overline I)||_1
\end{aligned}
\end{equation}
where  $\overline I$ denotes the target images and $L_{c}$ is the  Charbonnier loss with constant $\epsilon = 0.001$. $L_{e}$ is the edge loss, where $\triangle$ represents the  Laplacian operator. $L_{f}$  denotes the frequency domains loss, and $\mathcal{F}$ represents fast Fourier transform. To control the relative importance of loss terms, we set the parameters $\lambda = 0.1$ and $\delta = 0.05$  as in~\cite{Zamir2021MPRNet,FSNet}.

\section{More Ablation Studies}
\label{sec:mas}
We provide more ablation studies on the GoPro dataset~\cite{Gopro}.

\subsection{VBlock vs. MR-VNet~\cite{MR-VNet}}
Unlike MR-VNet~\cite{MR-VNet}, which treats the Volterra filter as a complete block, our VBlock incorporates it as part of a block. The VBlock first captures local features using LN, 1x1, and 3x3 convolutions before entering the Volterra filter, and then applies two 3x3 convolutions for each order branch. Additionally, by fusing the first- and second-order components and including a residual
connection for the preorder, our VBlock effectively avoids the issues of local feature loss. 

To validate the advantage of our VBlock over MR-VNet~\cite{MR-VNet}, we replace our VBlock with MR-VNet and present the results in Table 1. When replaced, the performance drops by 0.24 dB, demonstrating the superior performance of our VBlock. Additionally, MR-VNet~\cite{MR-VNet} is prone to gradient collapse during practical training.

\begin{table}
    \centering
       \caption{Results of alternatives to VBlock.}
    \label{tab:vbl1}
    \resizebox{\linewidth}{!}{
    \begin{tabular}{ccccc}
    \hline
         Net& MR-VNN~\cite{MR-VNet} &VBlock&  PSNR & $\triangle$ PSNR 
         \\
         \hline
         (a)& \ding{52} &   &34.51 & - 
         \\
         (b)& & \ding{52}   &34.75 & +0.24 
         \\
         \hline
    \end{tabular}}
\end{table}

\subsection{Resource Efficient}
We evaluate the model complexity of our proposed approach and other state-of-the-art methods in terms of running time and MACs. As shown in Table~\ref{tab:computational2222}, our method achieves the lowest MACs value while delivering competitive performance in terms of running time. 

\begin{table}
    \centering
    \caption{The evaluation of model computational complexity.}
    \label{tab:computational2222}
       \resizebox{\linewidth}{!}{
    \begin{tabular}{ccccc}
    \hline
         Method& Time(s) & MACs(G)  & PSNR$\uparrow$  & SSIM$\uparrow$ 
         \\
         \hline\hline
         MPRNet~\cite{Zamir2021MPRNet} & 1.148 & 777 & 32.66 &0.959
         \\
         Restormer~\cite{Zamir2021Restormer} & 1.218 & 140 & 32.92 & 0.961
         \\
         FSNet~\cite{FSNet} &\underline{0.362} & 111& 33.29 & 0.963
         \\
          MambaIR~\cite{guo2024mambair} &0.743 &439 &33.21 &0.962
         \\
         MR-VNet~\cite{MR-VNet} & 0.388 & \underline{96} & 34.04 & 0.969
         \\
         AdaRevD-L~\cite{AdaRevD} &0.761 & 460& \underline{34.60} & \underline{0.972}
         \\
         \hline
        \textbf{ DHNet(Ours)} &\textbf{0.256} &\textbf{32} & 33.28 & 0.964
         \\
        \textbf{DHNet-B(Ours)} & 0.499 &111 & \textbf{34.75} & \textbf{0.973}
         \\
         \hline
    \end{tabular}}
\end{table}

\section{Additional Visual Results}
\label{sec:Visual}
In this section, we present additional visual results alongside state-of-the-art methods to highlight the effectiveness of our proposed approach, as shown in Figures~\ref{fig:mgopr},\ref{fig:mhide},\ref{fig:mreal}.
It is clear that our model produces more visually appealing outputs for both synthetic and real-world motion deblurring compared to other methods.

\end{document}